\documentclass{Interspeech2024}

\interspeechcameraready

\title{Utilizing Multimodal Data for Edge Case Robust Call-sign Recognition and Understanding}

\name[]{Alexander}{Blatt}
\name[]{Dietrich}{Klakow}

\address{Saarland University, Saarland Informatics Campus, Germany}

\email{ablatt@lsv.uni-saarland.de}

\keywords{speech recognition, human-computer interaction, computational paralinguistics}
\keywords{robust, multimodal, NLU, ATC, safety}

\begin{document}

\maketitle

\begin{abstract}
Operational machine-learning based assistant systems must be robust in a wide range of scenarios. This hold especially true for the air-traffic control (ATC) domain. The robustness of an architecture is particularly evident in edge cases, such as high word error rate (WER) transcripts resulting from noisy ATC recordings or partial transcripts due to clipped recordings. To increase the edge-case robustness of call-sign recognition and understanding (CRU), a core tasks in ATC speech processing, we propose the multimodal call-sign-command recovery model (CCR). The CCR architecture leads to an increase in the edge case performance of up to 15\%.  We demonstrate this on our second proposed architecture, CallSBERT. A CRU model that has less parameters, can be fine-tuned noticeably faster and is more robust during fine-tuning than the state of the art for CRU. Furthermore, we demonstrate that optimizing for edge cases leads to a significantly higher accuracy across a wide operational range.
% The majority of natural language understanding (NLU) datasets are noise-free. Machine learning models trained on these datasets show a severe performance degradation and hallucinations, when applied to real-life data. This is particularly evident in edge cases, such as transcripts with high word error rates (WER), generated from noisy air-traffic control (ATC) recordings. In the worst case, the information extraction fails completely. We address this issue at the example of call-sign recognition and understanding (CRU), one of the core NLU tasks in ATC speech processing. We examine the three main edge cases that deteriorate the performance of CRU algorithms: \textit{high WER}, \textit{clipping} and \textit{missing transcript}. We introduce the call-sign-command recovery model (CCR) which exploits multimodal data to improve the edge case performance up to 15\% . Furthermore, we demonstrate that optimizing for edge cases leads to a significantly higher accuracy across a wide operational range.
\end{abstract}

\section{Introduction}
\label{sec:intro}
Pilots rely on the guidance of air-traffic controllers (ATCO) for a safe take-off and landing. Research projects targeting ACTO pilot communication automation, such as AcListant, Malorca \cite{Srinivasamurthy2018}, or ATCO2 \cite{proceedings2020059014}, are enabling the development of fully automated air traffic-control (ATC) speech processing pipelines and assistant systems. %The provision of prefiltered command choices for an ATCO \cite{Schmidt2014} or the automation of ATC speech collection for building an ATC data set \cite{Kocour2022} are building bricks towards full automation. Key speech-based applications for ATC are automatic speech recognition, automatic entity recognition, e.g. call-sign recognition, read-back error detection or stress monitoring, just to name a few. 
These systems should be robust and tested on edge case scenarios which the European Union Aviation Safety Agency states specifically \cite{EuropeanUnionAviationSafetyAgency2021}. This is contrasted by the fact that the majority of developed models are optimized for standard conditions on data sets like ATCOSIM \cite{Hofbauer2008},  AIRBUS \cite{Pellegrini2018} or ATCO2 \cite{Kocour2022}. These data sets are not designed for edge case testing and often lack edge case samples, like for example high noise recordings.
%Speech samples with a low SNR are, for example, filtered out, because they cannot be accurately transcribed by human annotators. 
This is problematic since high noise conditions with low SNR values are occurring during operation. If a machine learning (ML) system is not properly adapted to those conditions, hallucinations or drastic performance degradation can occur \cite{Ji2023}. %Additionally, for an increasing amount of models, statistical values like standard deviations or errors are not given, which is partially due to the big model sizes, that cause long training times that do not allow repetitions. Therefore, it is difficult for a user, e.g. an air navigation service provider (ANSP), to make reliable preoperational assumptions on the robustness and therefore safety aspects of the models that are trained and tested on standard ATC data sets. 
\par We address this at the example of call-sign recognition and understanding (CRU) \cite{Blatt2022}. Extracting the call-sign from ATC speech, respectively transcripts, is one of the key tasks in ATC. ACTOs address their commands to a specific pilot by starting each instruction with a call-sign\footnote{ATC  examples: https://wiki.flightgear.org/ATC\_ phraseology}. A misrecognized call-sign can lead to incidents or in the worst case accidents. Our first contribution to this topic is the introduction of CallSBERT, a novel, smaller and faster to train CRU model that can be used more flexible than the state of the art (SOTA) for CRU. As second contribution,  we show that training on edge cases like \textit{high WER}, \textit{clipping} and \textit{missing transcripts} can significantly improve the accuracy not only in these edge cases but over the whole operational range. We propose the call-sign-command recovery model (CCR) which utilizes commands and plane coordinates to recover additional call-sign accuracy (CA) in the edge cases and can even compensate for completely erroneous transcripts.

\section{Related work}
\label{sec:related work}
Related works focus on call-sign tagging \cite{Gupta2019a}, call-sign transcription \cite{Nigmatulina2021} or call-sign recognition in International Civil Aviation Organization (ICAO) format from ATC conversation transcripts \cite{Blatt2022,Ohneiser2021}. Multimodal approaches for automatic speech recognition (ASR) in ATC use surveillance call-signs from Automatic Dependent Surveillance–Broadcast (ADS-B) information to boost the performance  \cite{Guo2021,Kocour2022}. Blatt et al. propose a call-sign recognition and understanding (CRU) model using surveillance call-signs \cite{Blatt2022}. Their surveillance model, referred to as \textit{EncDec} model in the following, relies on ATC transcripts as input, which allows to evaluate the CRU task independently from the ASR task. This is the CRU reference model for our edge case optimization.  Plane locations are also useful context information, since commands are given from ATCOs to pilots usually at defined areas in the airspace. Kleinert et al. \cite{Kleinert2017} include plane locations via binary 2D airspace command distributions to improve their controller command prediction. We extend this idea, by using more informative non-binary 3D distributions in our command distribution module (CDM), which is one of the key components for our robust edge case CRU performance. Our CRU model CallSBERT is based on SBERT \cite{Reimers2019} and we adapt BERT \cite{Devlin2018a} as command classifier in our edge case robust CCR architecture.

\section{Data preparation}
\label{sec:Data preparation}
The CRU models are trained on ATC transcripts of the MALORCA data set (Prague airport) and on transcripts of the AIRBUS data set. Both data sets contain ATC transcripts labeled with the correct call-signs, e.g. \texttt{ryanair one two four} (expanded format), respectively \texttt{RYR124} (ICAO format) as shown in \autoref{fig:ccr}. The AIRBUS dataset, with artificial surveillance data added \cite{Blatt2022}, is only used for pretraining. This pretraining is crucial since the MALORCA dataset is relatively small. The train\textbar val\textbar test split consists of 0.9K\textbar 0.1K\textbar 0.1K samples for the MALORCA dataset, respectively 8.9K\textbar 1.3K\textbar 1.3K  samples for the AIRBUS dataset. To generate samples for command classification, the data is multi-labeled with a key-word-based labeler that recognizes six command types:  \textit{horizontal}, \textit{vertical}, \textit{ils}, \textit{taxi}, \textit{clearing} and \textit{greeting}. A transcript is tagged with \textit{horizontal} if it contains for example the key words \texttt{turn right} or \texttt{change heading}.\par 
For each of the MALORCA transcripts, the ADS-B information of each airplane in 100 km N-S and E-W distance of the Prague airport and 0-20 km altitude is fetched from the OpenSky data base\footnote{OpenSky: \url{https://opensky-network.org/}} via the timestamp of the transcript. From the ADS-B state vectors, the coordinates of the planes in the 200~km $\cdot$ 200~km $\cdot$ 20~km bounding box are isolated and transformed to an xyz coordinate system with its origin located at the airport. Roughly 30 planes are within this bounding box at the same time. Therefore a \textbf{random baseline} for call-sign identification has a chance of 1/30 to identify the correct call-sign.
\begin{comment}
\begin{figure}[t]
\begin{minipage}[b]{1.0\linewidth}
  \centering
  \centerline{\includegraphics[width=\linewidth]{figs/sample.pdf}}
%  \vspace{2.0cm}
\end{minipage}
\caption{Sample of the ADS-B enriched MALORCA data set. Samples of the AIRBUS data set are missing coordinates as input and command labels as output.}
\label{fig:sample}
%
\end{figure}
\end{comment}
For the different edge cases in \autoref{sec:WER}, \autoref{sec:Clipping} and \autoref{sec:empty}, the transcripts are altered accordingly. Versions of different WERs are produced by adding ASR noise as described in \cite{Blatt2022}. Additionally, clipped versions of the transcripts are produced by removing n words from the beginning of the transcript. All experiments are run on a NVIDIA GeForce RTX 2060 GPU. All experiments are run thrice and the mean and standard deviation are given. For each run, the model with the lowest validation loss is chosen for testing.

\begin{figure}[t]
\begin{minipage}[b]{1.0\linewidth}
 \centering
  \centerline{\includegraphics[width=0.8\linewidth]{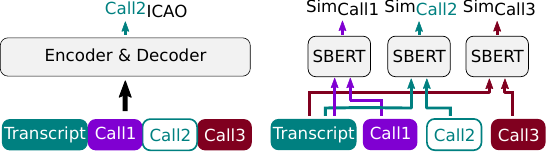}}
  \vspace{0.2cm}
\end{minipage}
\caption{Architecture comparison of the parallel EncDec \cite{Blatt2022} (left)  and the sequential CallSBERT model (right).}
\label{fig:archComp}
\end{figure}
\section{Models}
\label{sec:models}

\subsection{EncDec}
As SOTA, we take the EncDec model from Blatt et al. \cite{Blatt2022} which uses a \textit{bert-base}\footnote{Transformer library: \url{https://huggingface.co}} encoder-decoder architecture and has 66.3M parameters. One mayor drawback of the EncDec architecture is the way, the model is trained. The model input consists of a transcript concatenated with \textbf{all} surveillance call-signs to predict the target call-sign directly in ICAO format as \autoref{fig:archComp}a shows.
\subsection{CallSBERT}
The CallSBERT model takes the transcript and only \textbf{one} matching or non-matching surveillance call-sign for the contrastive loss training. This significantly reduces the input size. In \autoref{fig:archComp}, \textit{Call2} is an example for a matching call-sign (positive sample), while \textit{Call1} is a non-matching call-sign (negative sample).  The CallSBERT architecture is based one SBERT block\footnote{SBERT library: \url{https://www.sbert.net}} \cite{Reimers2019}, visualized in \autoref{fig:archComp}, and has only 37.1\% (24.6M parameters) of the EncDec model parameters. All this results in an increased training speed of a \textbf{factor of 4}\footnote{roughly 100~s vs 400~s for 10 epochs finetuning on the 0.9K MALORCA train split} in comparison with the EncDec Model. If the models are applied to a bigger airspace, with more surveillance call-signs present, this factor will further increase. Since CallSBERT ranks the surveillance call-signs \textbf{sequentially} during inference via cosine-similarity scores (\textit{Sim}), its maximum input size does not need to be defined beforehand, which is an advantage of this architecture. The production of similarity scores also allows this architecture to be used in a submodel, because the similarity scores for each surveillance call-sign can be used as features. In contrast, the EncDec architecture does only predict one call-sign and gives no information about the other call-signs which is the main disadvantage of this model.
\begin{figure}[t]
\begin{minipage}[b]{1.0\linewidth}
  \centering
  \centerline{\includegraphics[width=\linewidth]{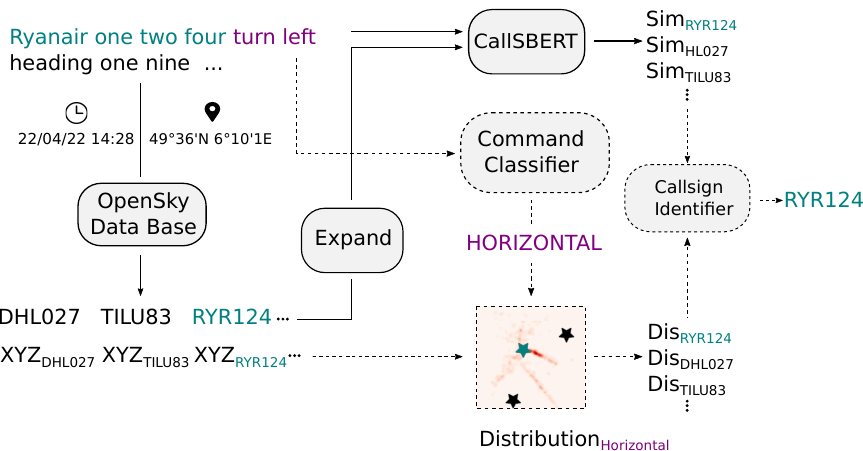}}
\end{minipage}
\caption{CCR architecture. The dotted lines mark the additional call-sign prediction path via command distributions.}
\label{fig:ccr}
\end{figure}
\subsection{CCR}
The call-sign-command recovery model (CCR), displayed in \autoref{fig:ccr}, combines command with call-sign recognition to increase the robustness of the CRU. It consists of a CallSBERT branch (solid lines) and the additional command branch (dotted lines), which utilizes coordinates as additional input. The command branch consists of three different modules, the command classifier, the command distribution module (CDM) and the final call-sign identifier.
\begin{comment}   
\begin{table}[ht]
\centering
\caption{Model Size Comparison}
\label{tab:size}
\begin{tabular}[t]{lcc}
\toprule
&EncDec &CallSBERT\\
\midrule
Parameter&66.362.880&24.617.794\\
Relative&100\%&37.1\%\\
\bottomrule
\end{tabular}
\end{table}
\end{comment}
The command classifier is a transformer-based multi-label classifier. It can detect whether a transcript contains one or multiple of the six command types described in \autoref{sec:Data preparation}. The predicted command types are  fed into the command distribution module (CDM). The CDM consists of plane 2D/3D-coordinates \textrightarrow~command probabilities (\textit{Dis}) mappings for each of the six command types. The CDM contains mappings for each command type and they are selected based on the command types that are recognized by the command classifier. The \textit{Dis} scores therefore indicate which plane in the airspace is most likely mentioned in the transcript based on its position and the command uttered in the transcript. In the example in \autoref{fig:ccr}, just the \textit{horizontal} command type is identified, therefore the CDM only uses the probability distribution map of the \textit{horizontal} command for the \textit{Dis} generation.
\begin{figure}[t]
\begin{minipage}[b]{.49\linewidth}
  \centering
  \centerline{\includegraphics[width=\linewidth]{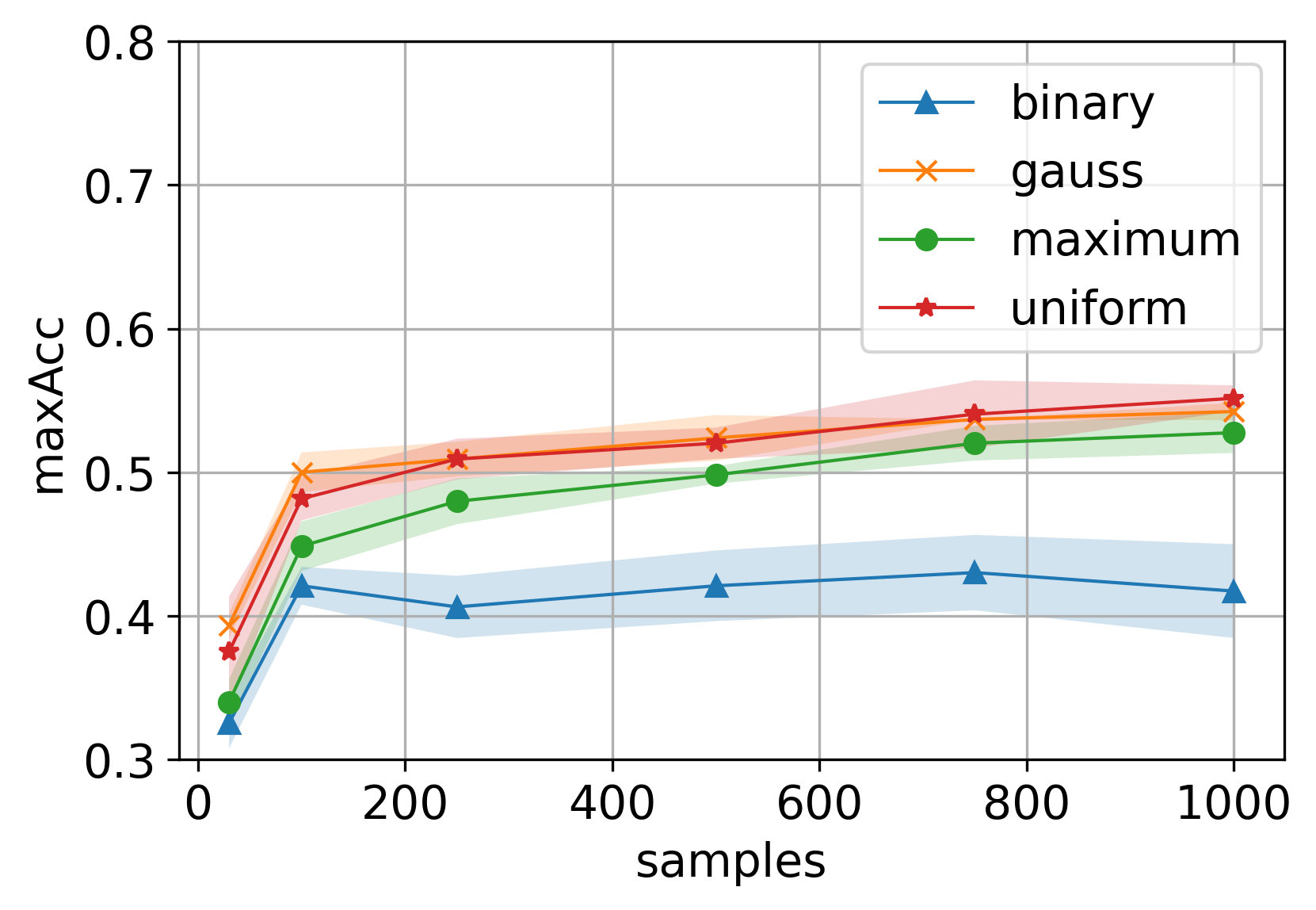}}
%  \vspace{1.5cm}
  \centerline{(a) 2D select}\medskip
\end{minipage}
\hfill
\begin{minipage}[b]{0.49\linewidth}
  \centering
  \centerline{\includegraphics[width=\linewidth]{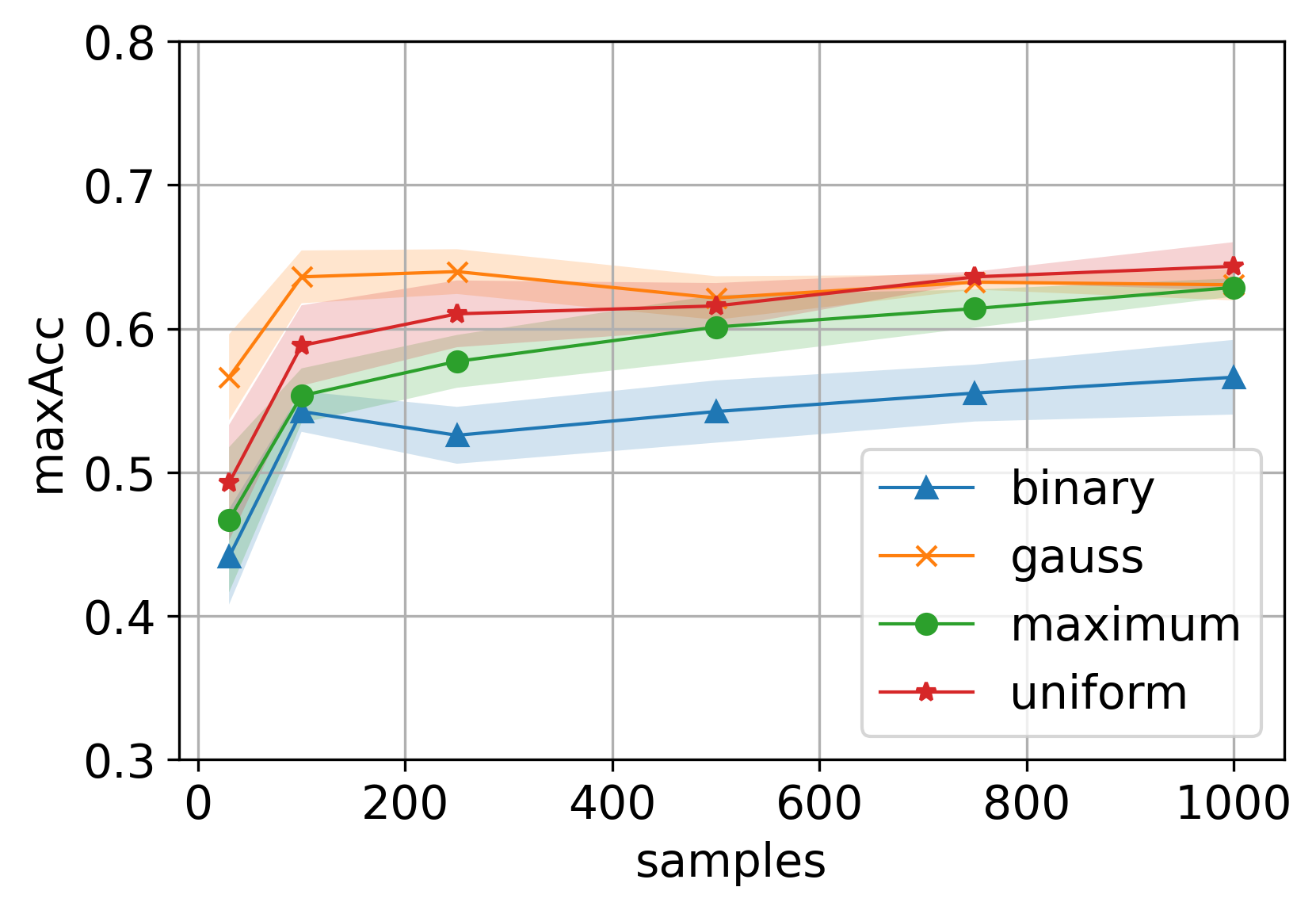}}
%  \vspace{1.5cm}
  \centerline{(b) 3D select}\medskip
\end{minipage}
\begin{minipage}[b]{.49\linewidth}
  \centering
  \centerline{\includegraphics[width=\linewidth]{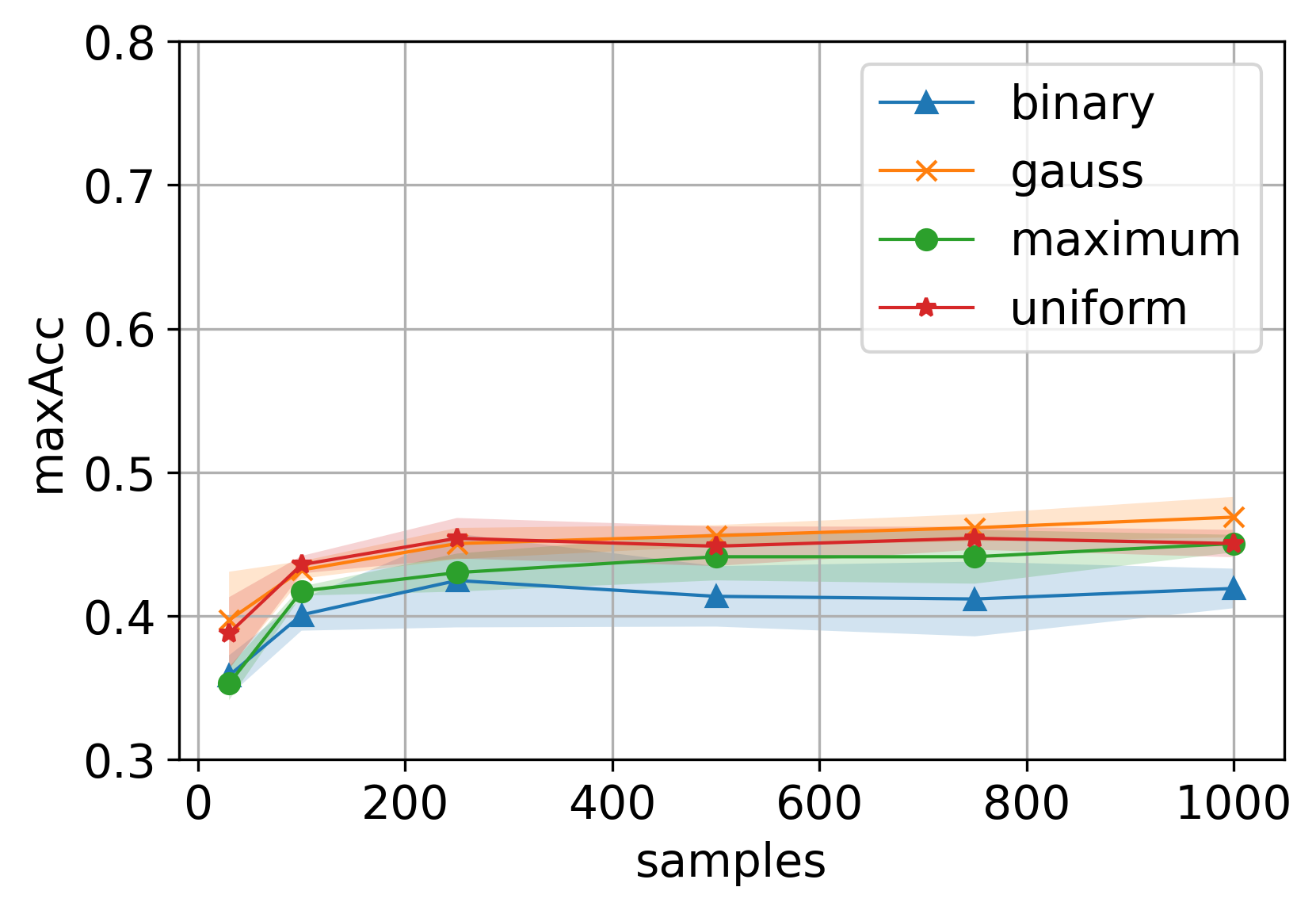}}
%  \vspace{1.5cm}
  \centerline{(c) 2D naive}\medskip
\end{minipage}
\hfill
\begin{minipage}[b]{0.49\linewidth}
  \centering
  \centerline{\includegraphics[width=\linewidth]{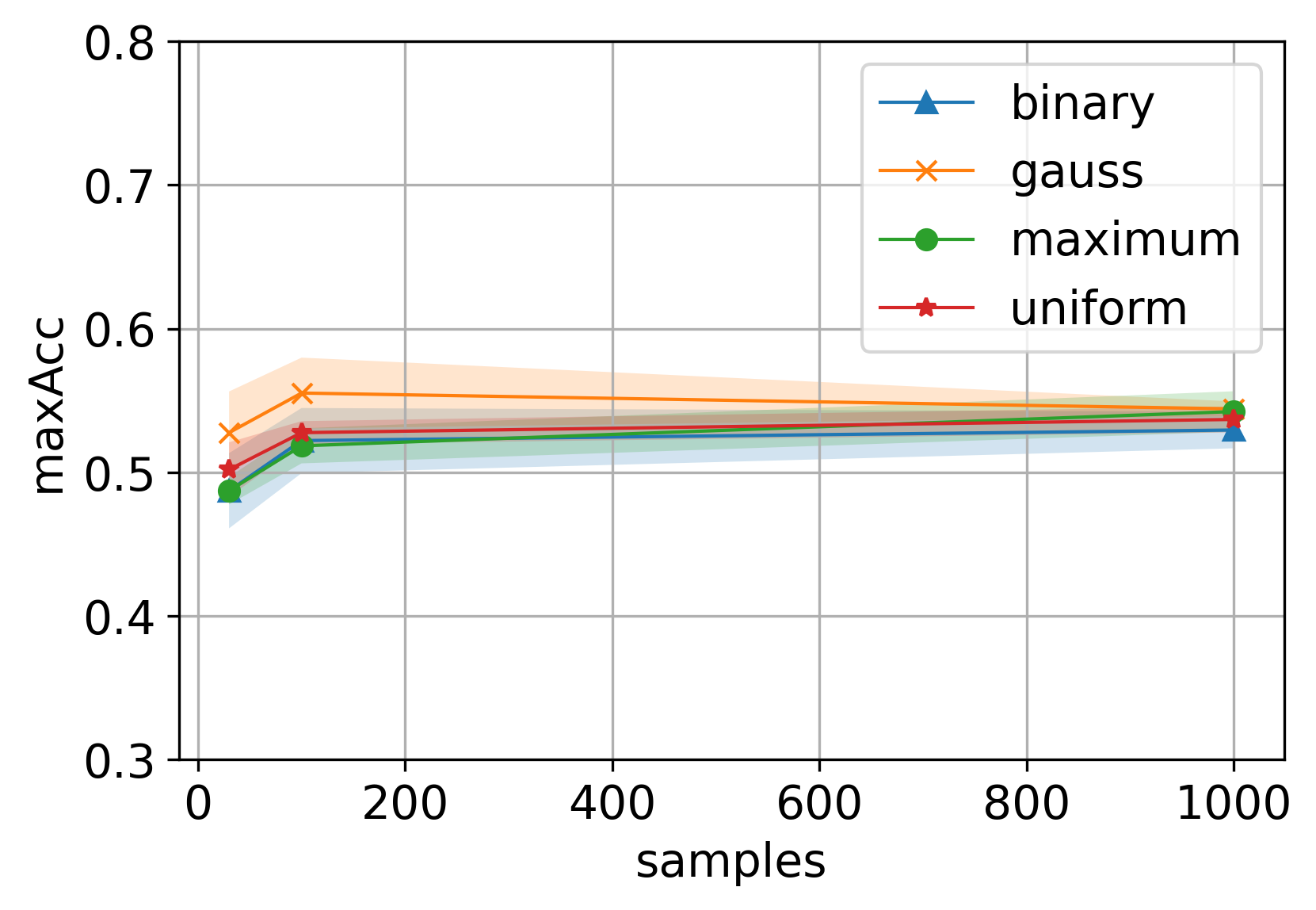}}
%  \vspace{1.5cm}
  \centerline{(d) 3D naive}\medskip
\end{minipage}
\caption{Maximum accuracy of call-sign prediction based on command distributions with optimal filter parameters.}
\vspace{-2mm}
\label{fig:filterSelect}
\end{figure}

\begin{figure}[b]
\vspace{-3 mm} 
\begin{minipage}[b]{.32\linewidth}
  \centering
  \centerline{\includegraphics[width=\linewidth,trim={2cm 0 2.3cm 1.1cm},clip]{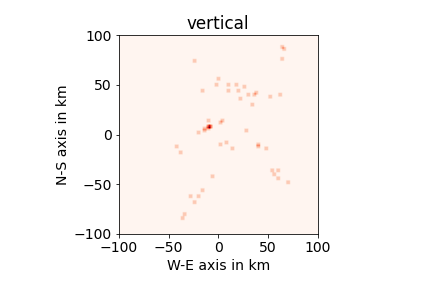}}
%  \vspace{1.5cm}
  \centerline{(a) Plane coordinates}\medskip
\end{minipage}
\hfill
\begin{minipage}[b]{0.32\linewidth}
  \centering
  \centerline{\includegraphics[width=\linewidth,trim={2cm 0 2.3cm 1.1cm},clip]{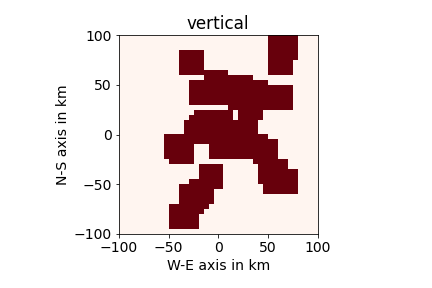}}
%  \vspace{1.5cm}
  \centerline{(b) Binary filtered}\medskip
\end{minipage}
\hfill
\begin{minipage}[b]{0.32 \linewidth}
  \centering
  \centerline{\includegraphics[width=\linewidth,trim={2cm 0 2.3cm 1.1cm},clip]{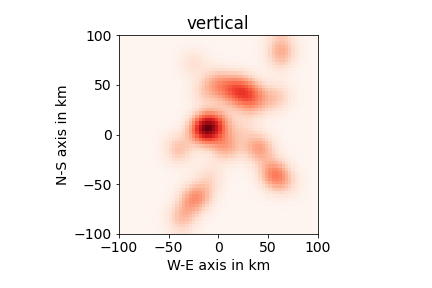}}
%  \vspace{1.5cm}
  \centerline{(c) Gaussian filtered}\medskip
\end{minipage}
\caption{ 2D coordinates of airplanes while receiving a vertical command (a) and 2D distribution maps (top view) of the vertical command in the 200~km $\cdot$ 200~km Prague airspace (b),(d). Dark colored areas have a high probability for vertical commands.}
\vspace{3 mm}
\label{fig:heatmaps}
\end{figure}If there is no transcript available, the coordinate \textrightarrow~probability mappings for every command type are considered and mean pooled. For generating the mappings, a small set of coordinate-command pairs of the target airspace are filtered by one of the following filter functions: \textit{Gaussian}, \textit{binary}, \textit{maximum} or \textit{uniform}. The filtering, described in \autoref{sec:Command Distributions} allows to generate command probability distributions for the whole airspace out of just a few hundred samples as \autoref{fig:heatmaps} shows. The final call-sign identifier of the CCR model takes the \textit{Sim} scores of CallSBERT and the \textit{Dis} scores of the CDM module for each surveillance call-sign and generates a final weighted score for each surveillance call-sign and extracts the most probable one.  Our identifier consists of a fully connected five-layer network with relu activations and batch normalization in between the fully connected layers and a sigmoid activation at the last layer. 

\subsection{CDM optimization}
\label{sec:Command Distributions}
To reduce the need for a large transcribed corpus to create the command probability distributions of the CDM for a new airspace, we evaluate different filter functions for the distribution generation in a low-resource scenario. The naive baseline uses all command distributions for the call-sign prediction (\textit{naive mode}). Using just the relevant command distributions (\textit{select mode}), which are selected by the command classifier of the CCR, adds 10\% accuracy over the naive baseline for the \textit{Gaussian}, \textit{maximum} and \textit{uniform} filter as \autoref{fig:filterSelect} shows. Switching from of 2D coordinates to 3D coordinates, respectively incorporating the plane height, additionally adds 10\% accuracy. The highest accuracy for the low resource scenario is achieved with a Gaussian filter. Using just 100  coordinate-command pairs to generate the distributions  via Gaussian filtering gives a similar accuracy as with 1000 samples for the \textit{3D select} case. For the final CCR model, we therefore use 3D coordinate\textrightarrow probability mappings generated by a Gaussian filtering as shown in \autoref{fig:heatmaps}c.

\begin{figure}[t]
\begin{minipage}[b]{.48\linewidth}
  \centering
  \centerline{\includegraphics[width=\linewidth]{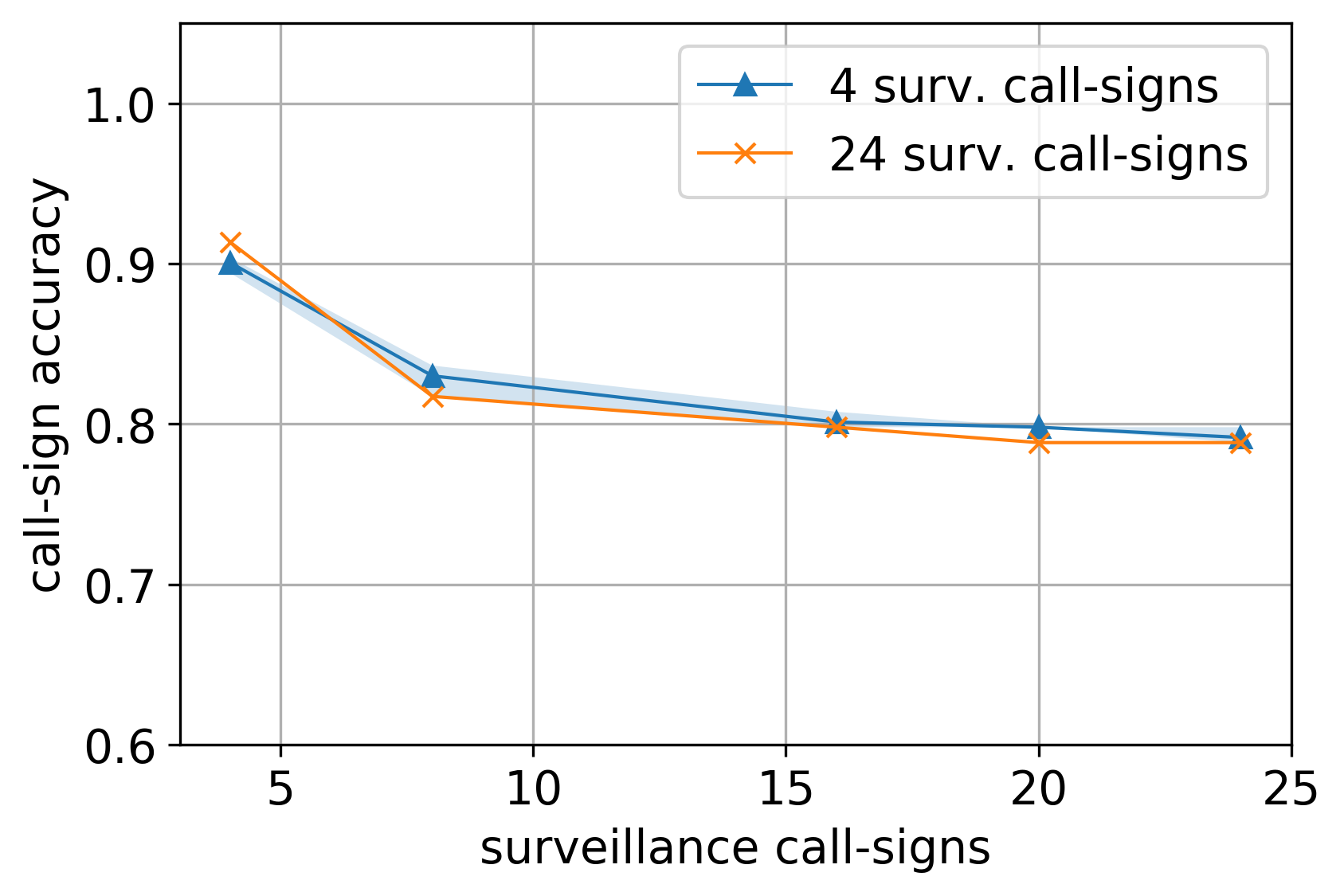}}
%  \vspace{1.5cm}
  \centerline{(a) CallSBERT }\medskip
\end{minipage}
\hfill
\begin{minipage}[b]{0.48\linewidth}
  \centering
  \centerline{\includegraphics[width=\linewidth]{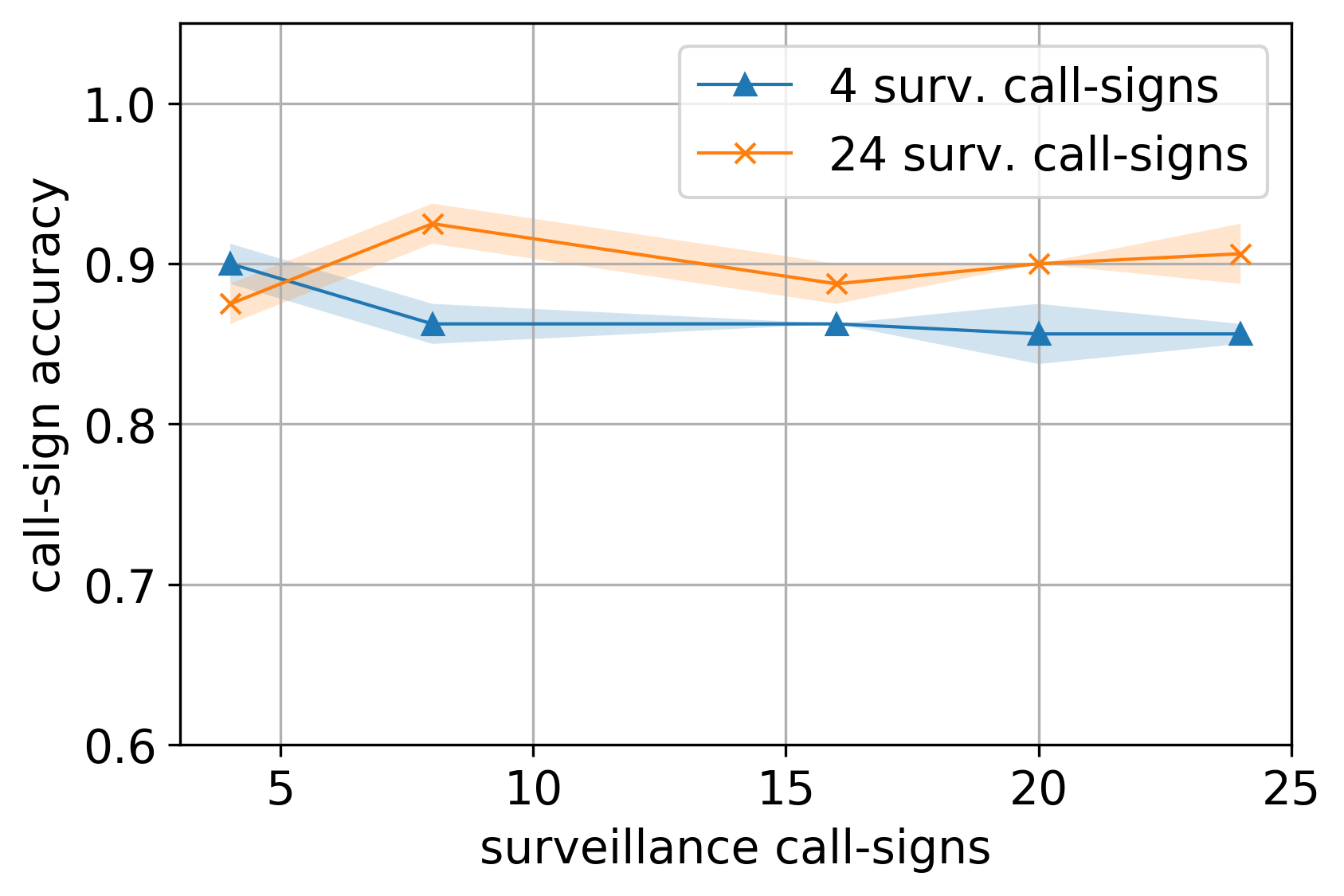}}
%  \vspace{1.5cm}
  \centerline{(b) EncDec }\medskip
\end{minipage}
\caption{Call-sign accuracy depending on the surveillance size per test transcript. During fine-tuning, each transcript has either 4 or 24 corresponding surveillance call-signs.}
\label{fig:callsign} 
\end{figure}
\section{Results}
\label{sec:Results}

\subsection{CallSBERT: Surveillance adaptation}
\label{sec:Surveillane adaptation}

Depending on the the flight sector, the amount of surveillance call-signs available might vary. The EncDec architecture is proven to be robust against fluctuations in the surveillance call-sign count during testing \cite{Blatt2022}. The question remains how the EncDec architecture and CallSBERT react, when they are fine-tuned with a different amount of surveillance call-signs. \autoref{fig:callsign}b shows that the CA of the EncDec model, despite staying over 80\%, depends on the number of surveillance call-signs encountered during training. If the model is finetuned on samples with 24 surveillance call-signs per transcript, it performs better if the number of surveillance call-signs during testing is in the same range. The same holds true for the model trained with 4 surveillance call-signs per transcript. The CallSBERT model however seems to be agnostic against the number of surveillance call-signs encountered during training and shows the expected behaviour of a reduced CA with an increasing number of surveillance call-signs due to an increasing search-space. 

\subsection{Edge cases}
\subsubsection{High word error rate}
\label{sec:WER}
\begin{figure}[t]
% \begin{minipage}[b]{.48\linewidth}
%   \centering
%   \centerline{\includegraphics[width=\linewidth]{figs/SBERTAirAir.png}}
% %  \vspace{1.5cm}
%   \centerline{(a) CallSBERT Airbus}\medskip
% \end{minipage}
% \hfill
% \begin{minipage}[b]{0.48\linewidth}
%   \centering
%   \centerline{\includegraphics[width=\linewidth]{figs/EncDecAirAir.png}}
% %  \vspace{1.5cm}
%   \centerline{(b) EncDec Airbus}\medskip
% \end{minipage}
%
\begin{minipage}[b]{.49\linewidth}
  \centering
  \centerline{\includegraphics[width=\linewidth]{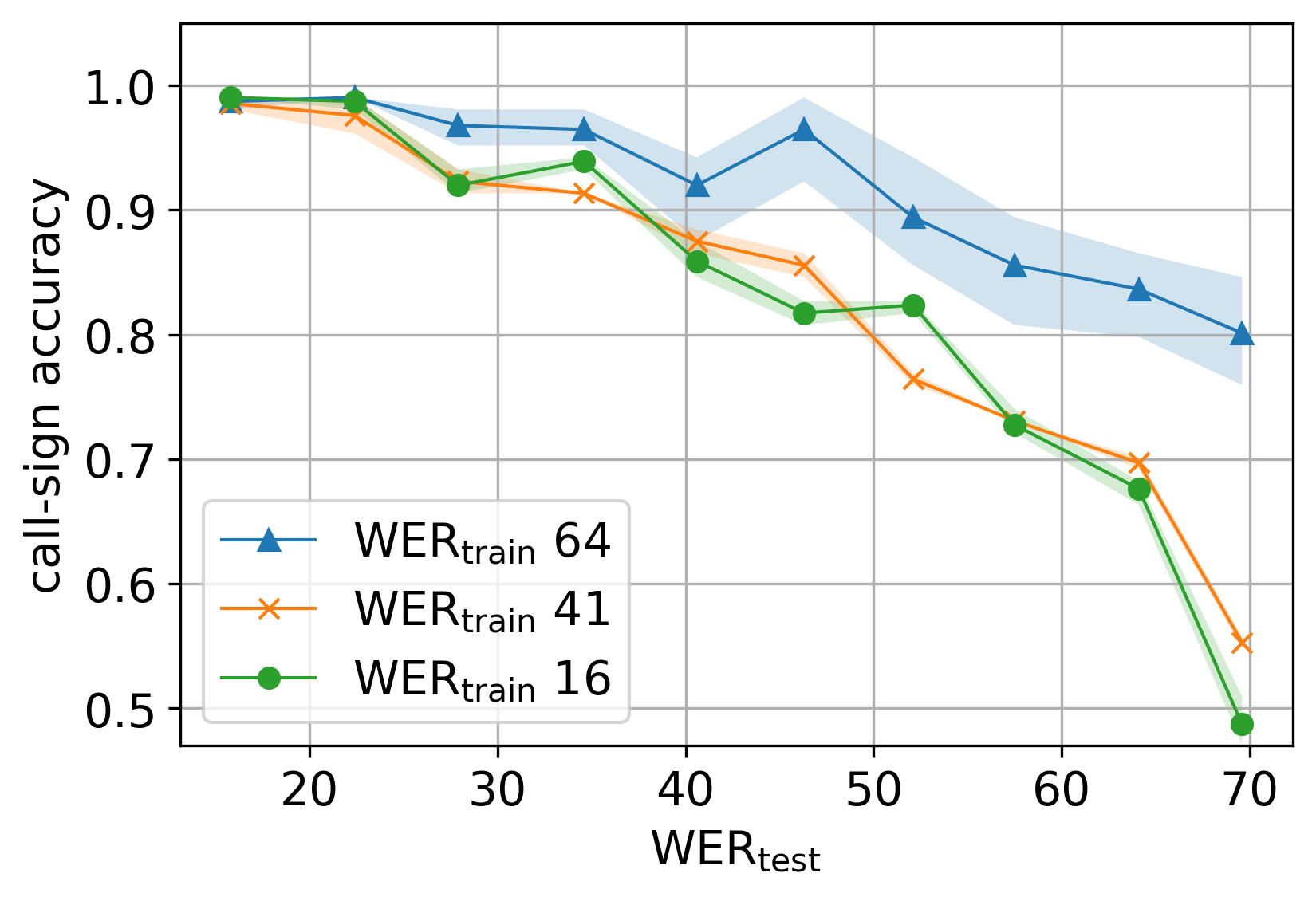}}
%  \vspace{1.5cm}
  \centerline{(a) CallSBERT}\medskip
\end{minipage}
\hfill
\begin{minipage}[b]{0.49\linewidth}
  \centering
  \centerline{\includegraphics[width=\linewidth]{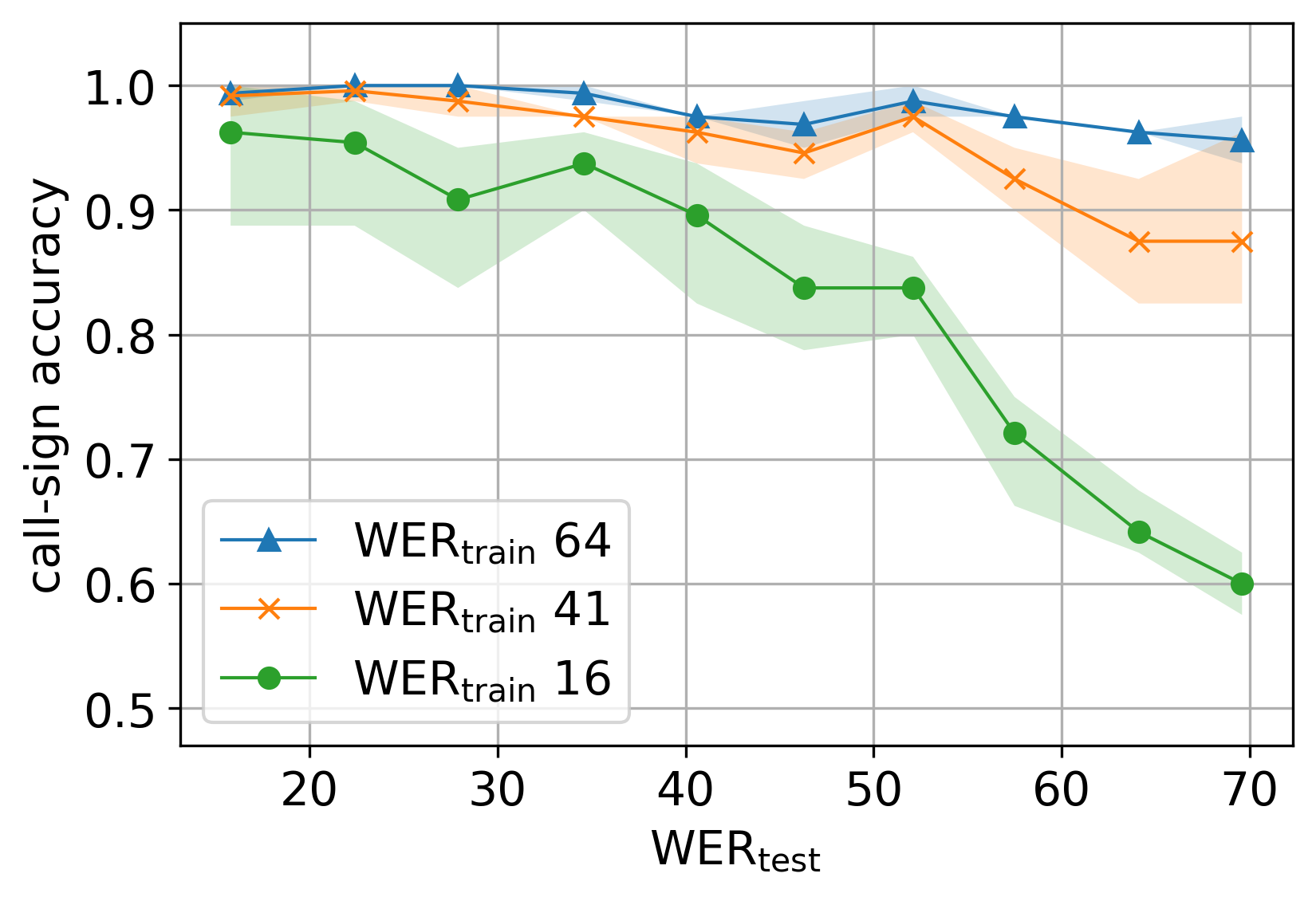}}
%  \vspace{1.5cm}
  \centerline{(b) EncDec}\medskip
\end{minipage}
\hfill
\begin{minipage}[b]{0.49\linewidth}
  \centering
  \centerline{\includegraphics[width=\linewidth]{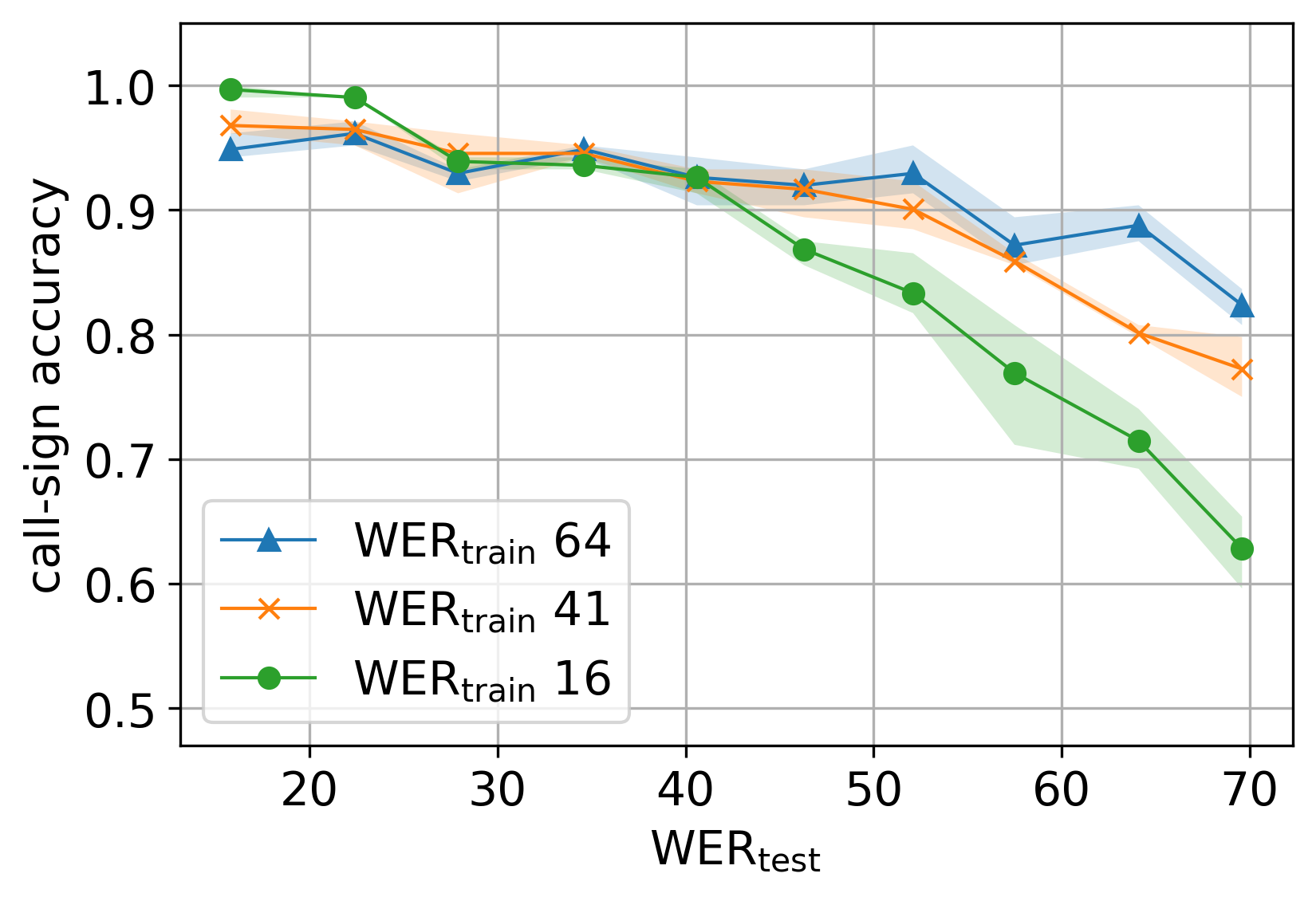}}
  %\vspace{0.32cm}
  \centerline{(c) CCR}\medskip
\end{minipage}
\hfill
\begin{minipage}[b]{0.49\linewidth}
  \centering
  \centerline{\includegraphics[width=\linewidth]{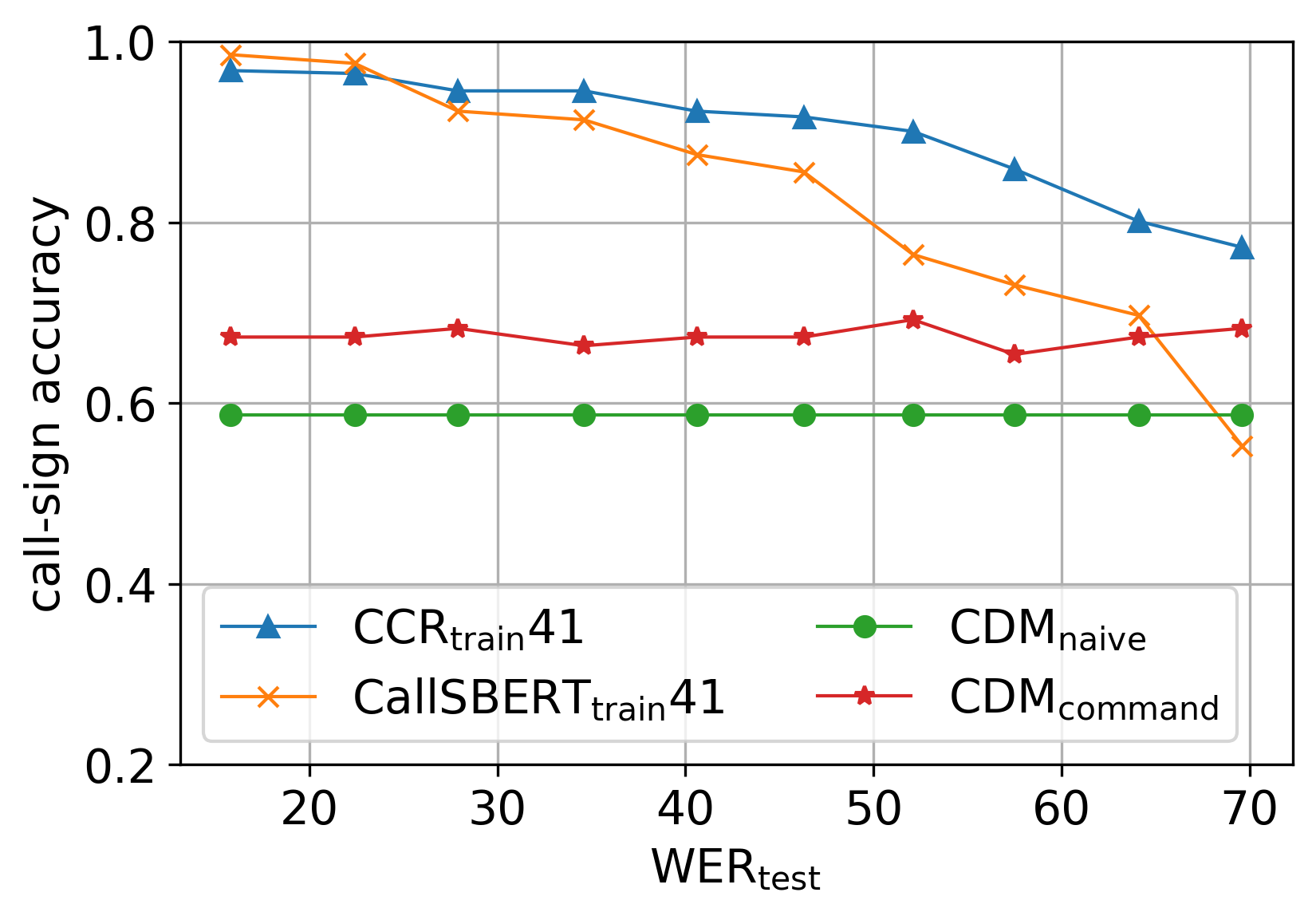}}
%  \vspace{1.5cm}
  \centerline{(d) CCR ablation study}\medskip
\end{minipage}
\caption{Call-sign accuracy depending on the WER of the MALORCA test data.}
\label{fig:WER}
\end{figure}

The best performing SOTA ASR model of \cite{Kocour2022} achieves a mean word error rate (WER) on their LiveATC data set \cite{Kocour2022} of 26.8\%. But we found that 24\% of the transcripts have a 40\% WER or higher and 9\% of the transcripts have even a WER over 60\%. For our experiments, we therefore generate test data sets with a mean WER of up to 70\%. \autoref{fig:WER} shows that both, the EncDec and CallSBERT model show a significant performance drop at high WERs, when trained on low WER data of 16\%. Training on data with higher WERs allows the models to learn the noise distribution and reduces the CA deterioration by up to 30\%, with the bigger EncDec model adapting  better to the ASR noise. Incorporating the CallSBERT model into the CCR architecture stabilizes the CA for a WER over 60\% and adds up to 15\% to the accuracy of the pure CallSBERT model. %This accuracy increase is purely based on the additional information the model draws from the plane coordinates and the command extracted from the transcript. This suggests, that the high resistance of the CCR command classifier against high WERs, as shown in  \autoref{fig:WER}(f), is probably a key factor.
\begin{comment}
\begin{figure}[htb]
\begin{minipage}[b]{1.0\linewidth}
  \centering
  \centerline{\includegraphics[width=0.5 \linewidth]{figs/ablation.png}}
%  \vspace{2.0cm}
\end{minipage}
\caption{Ablation study of the CCR module.}
\label{fig:ablation}
\end{figure}
\end{comment}
\par
To further evaluate this, we conduct an ablation study on the CCR architecture. In the CDM\textsubscript{naive} case, the CDM mean pools the output of all 3D command distributions to generate a score for each plane coordinate. Since this part of the CCR is not depending on ASR output, the accuracy is stable over the whole WER range as \autoref{fig:WER}d shows. By feeding the output of the command classifier into the CDM (CDM\textsubscript{command}), the CDM selects the distribution map of the most probable command for predicting the call-sign. This adds roughly 10\% performance as \autoref{fig:WER}d shows. The missing deterioration of the accuracy at high WERs proves the robustness of the command prediction. Up to a test WER of 40\%, the call-sign accuracy of CallSBERTs is more than 20\% higher than the CA of CDM\textsubscript{command}. At higher WERs the accuracy of CallSBERT drops significantly. The full CCR architecture however outperforms the single CCR modules by combining the CDM\textsubscript{command} and CallSBERTs output. The ablation study highlights the importance of using multi-modal data, but also the importance of extracting noise-robust text-based features. 
\begin{figure}[b]
\begin{minipage}[b]{.48\linewidth}
  \centering
  \centerline{\includegraphics[width=\linewidth]{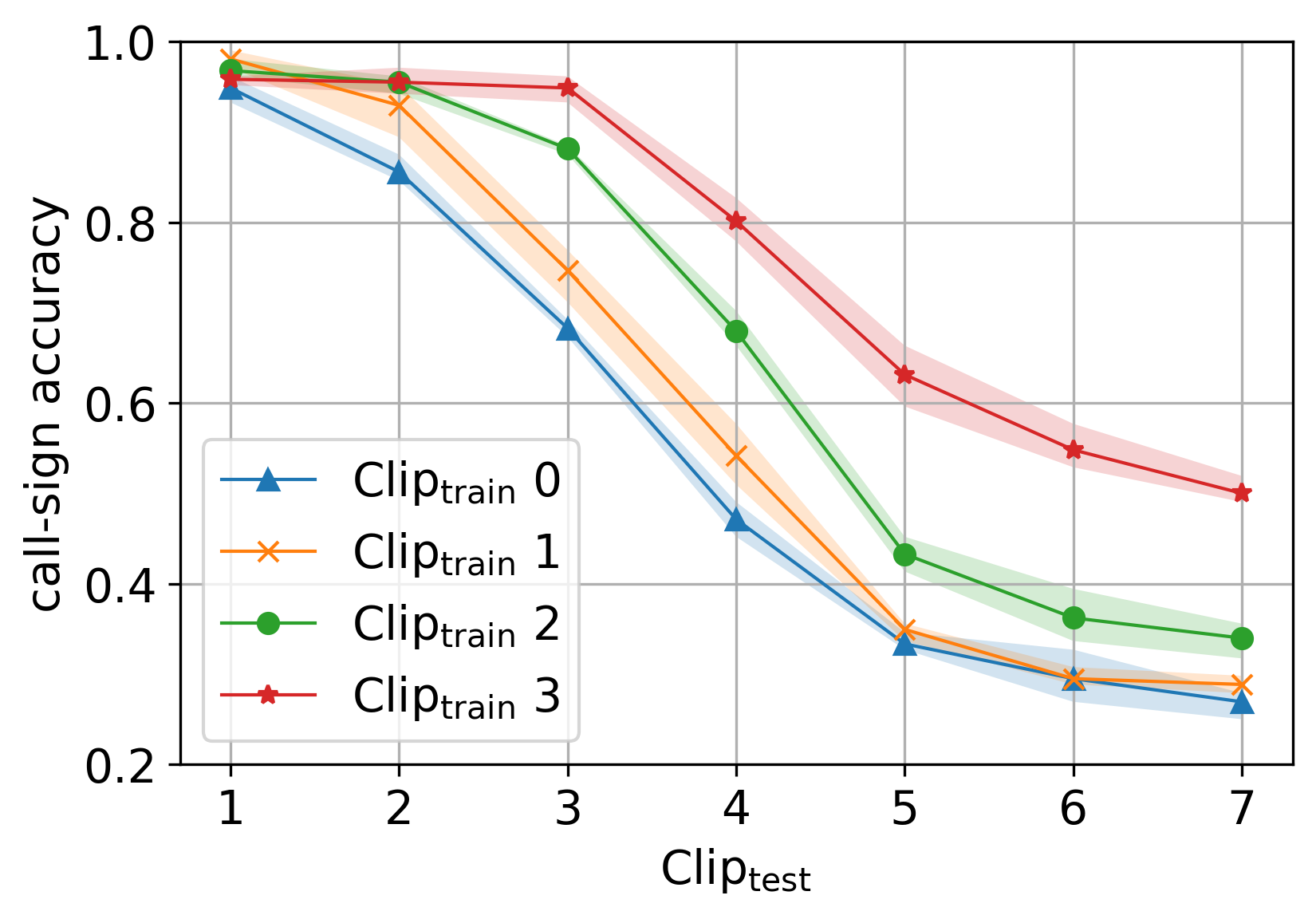}}
%  \vspace{1.5cm}
  \centerline{(a) CallSBERT }\medskip
\end{minipage}
\hfill
\begin{minipage}[b]{0.48\linewidth}
  \centering
  \centerline{\includegraphics[width=\linewidth]{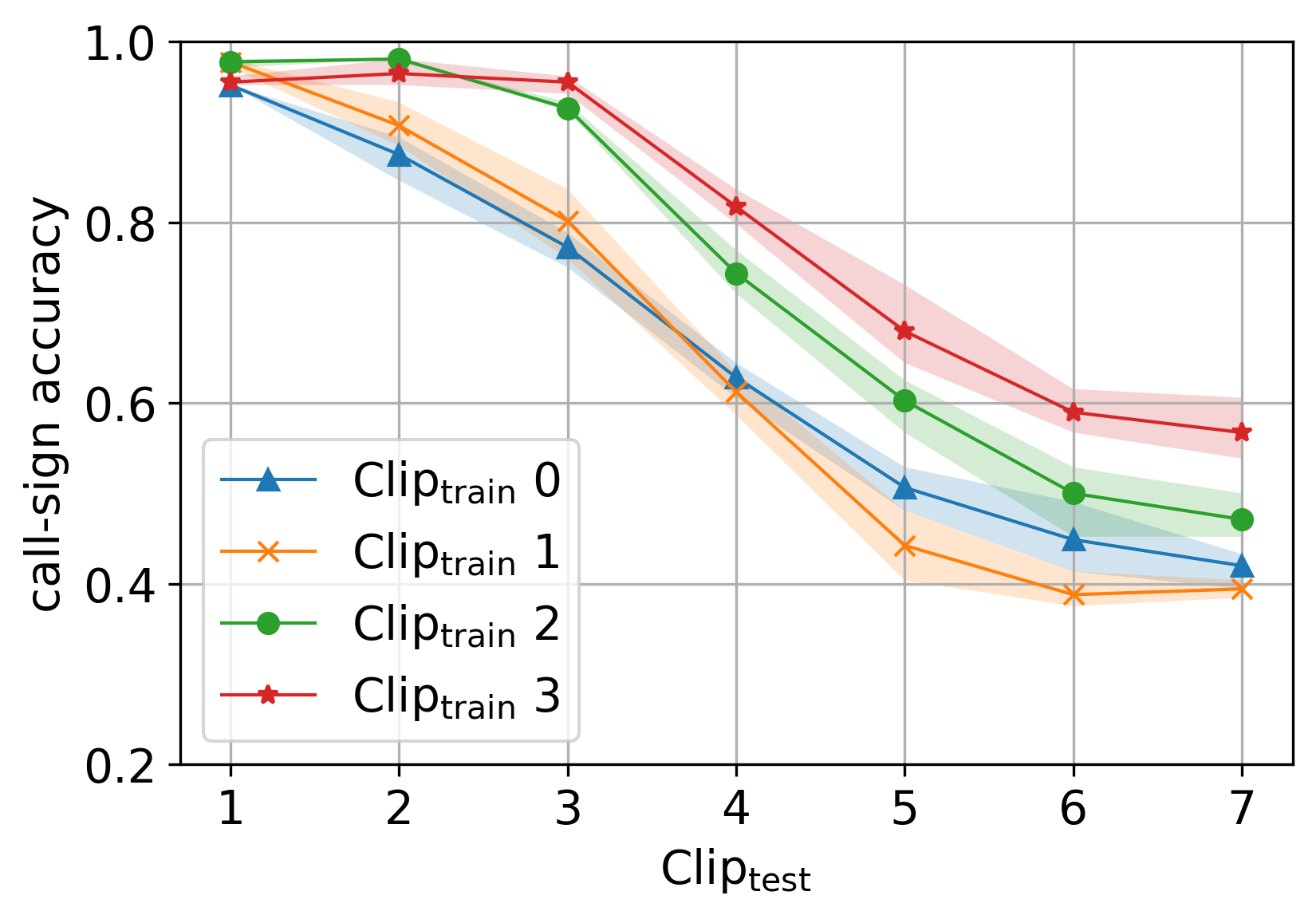}}
%  \vspace{1.5cm}
  \centerline{(b) CCR}\medskip
\end{minipage}
\begin{minipage}[b]{0.48\linewidth}
  \centering
  \centerline{\includegraphics[width=\linewidth]{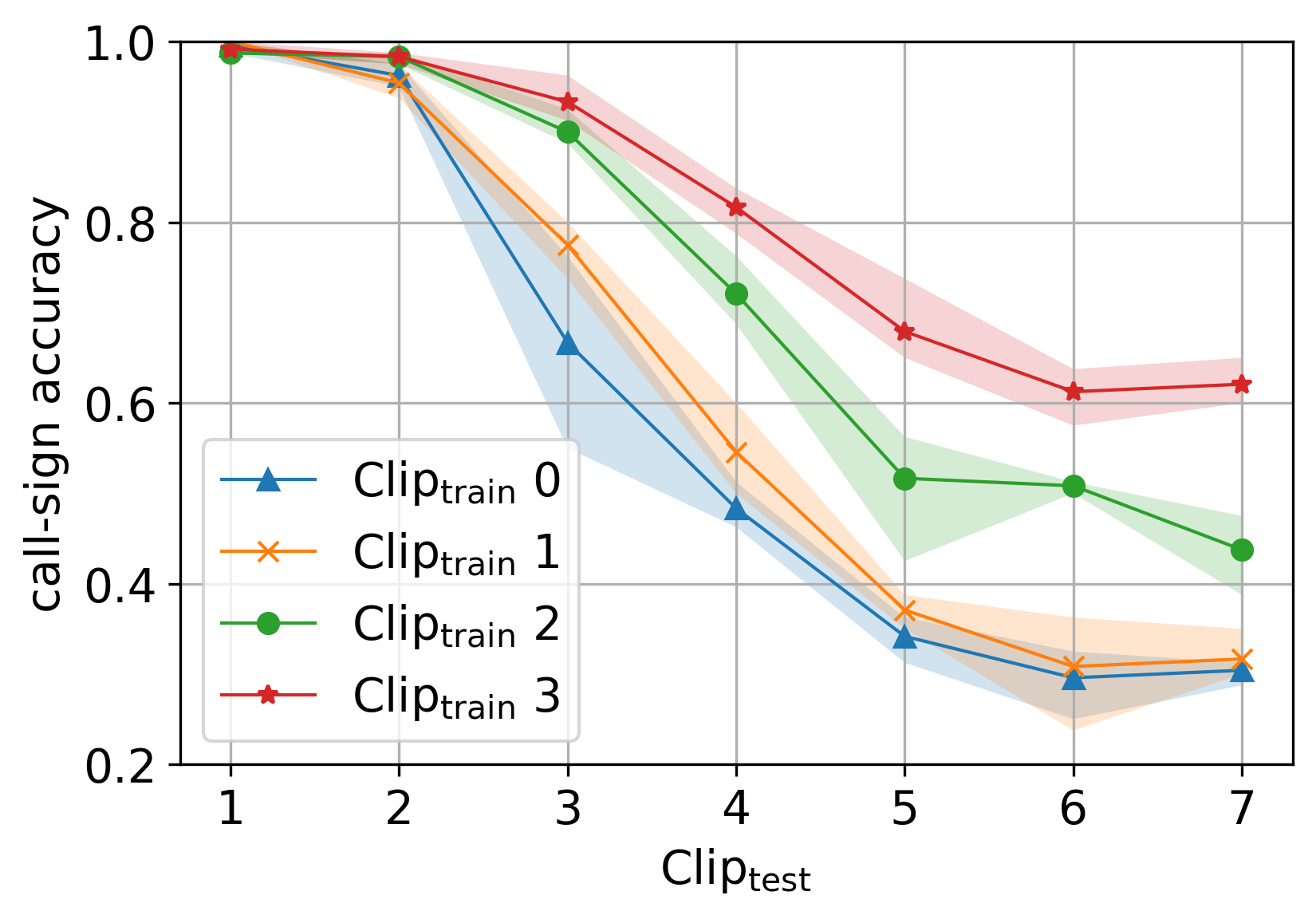}}
%  \vspace{1.5cm}
  \centerline{(c) EncDec}\medskip
\end{minipage}
\hfill
\begin{minipage}[b]{0.44\linewidth}
  \caption{Call-sign accuracy depending on the number of words clipped off at the beginning of the transcripts.}
  \label{fig:Clip}
  \vspace{25pt}
\end{minipage}

\end{figure}

\subsubsection{Clipping}
\label{sec:Clipping}
An ATC utterance can be clipped at the beginning if the transmission of the utterance starts delayed after the ATCO or pilot started talking. CRU algorithms are quite sensitive to clipping, since the call-sign is either located at the beginning or end of an utterance. Clipping for example the first three words of the call-signs \texttt{lufthansa one two four lima echo} and \texttt{ryanair three five four lima echo} results in the identical call-sign. \autoref{fig:Clip} shows that clipping just the first four words reduces the CA of CallSBERT below 50\%, which compares to a WER higher than 70\%. With the clipping of six words, the CA starts to plateau, since the majority of call-signs at the beginning of utterances are cut off beyond recognition. Training specifically on those shortened utterances can recover up to 30\% CA. The additional command branch of the CCR reduces the performance drop by 10\%, even for an CallSBERT model, which is trained on unclipped data. The comparison between the CCR module and the EncDec architecture shows that both architectures have a similar performance, when they encounter already heavily clipped data during training. If however only one or no words are clipped during training the CCR architecture outperforms the EncDec model significantly.

\subsubsection{Missing transcript}
\label{sec:empty}
The worst case scenario for a CRU model is a missing transcript. In the ATCO² project, utterances with an SNR $<$ 0~dB, make up roughly 10\% of all the recordings. They are however discarded because they are to noisy for ASR. To still make use of such samples, a CRU model has to work solely on surveillance data. \autoref{tab:empty} shows, that the CallSBERT model, cannot utilize the surveillance call-signs to reach a CA higher than 10\% if the transcript is completely missing. The EncDec model is able to generate predictions, when trained on the 64\% WER data because the model utilize the simultaneous processing of all surveillance call-signs to draw a prediction from previous surveillance constellations. It falls however far behind the CCR model for lower WER training data and fails completely at 16\% WER training data. The additional command distribution maps keeps the CCR module still operational at 16\% WER, where the other CRU models completely break down. 

\begin{table}[tb]
\footnotesize
\caption{Call-sign accuracy on test data without transcripts.}
\label{tab:empty}
\centering
\begin{tabular}[t]{lccc}
\toprule
WER\textsubscript{train}&16\% &41\% & 64\%\\
\midrule
CallSBERT&0.03($\pm$0.02)&0.07($\pm$0.06)&0.06($\pm$0.04)\\
EncDec&0.00($\pm$0.00)&0.12($\pm$0.04)&0.31($\pm$0.05)\\
CCR&\textbf{0.16}($\pm$0.04)&\textbf{0.33($\pm$0.03)}&\textbf{0.37}($\pm$0.04)\\

\bottomrule
\end{tabular}

\end{table}

\section{Conclusion}
\label{sec:summary}
In this work we have shown at the example of call-sign recognition and understanding models, that edge case optimization leads to a more stable performance over a broad operational range. Fine-tuning on noisy transcripts reduces the noise introduced accuracy drop significantly without degrading accuracy levels on clean data. This holds true for high WER transcripts as well as for word-clipped trancripts. Our introduced CallSBERT model shows just a minor performance decrease compared to the EncDec model introduced in \cite{Blatt2022} while having only 37.1\% of the parameters and beeing faster and more robust during finetuning. This performance gap is significantly reduced when CallSBERT is integrated in our newly proposed multimodal CCR architecture. The ablation study of the architecture shows, that the additional context information extracted by the command distribution module and the command classification module of the CCR architecture ensures a stable performance for all investigated edge case scenarios. This makes this design also interesting for other domains, where coordinates of communication targets are known, for example the nautical or the military domain. Due to its command distribution module, the CCR model can even produce nearly 40\% accurate predictions when there is no transcript available, making it the favorable choice for a robust call-sign prediction model.

\bibliographystyle{IEEEtran}
\bibliography{mybib}

\end{document}